\documentclass[conference]{IEEEtran}
\IEEEoverridecommandlockouts
\usepackage{cite}
\usepackage{amsmath,amssymb,amsfonts}
\usepackage{algorithmic}
\usepackage{graphicx}
\usepackage{textcomp}
\usepackage{xcolor}
\usepackage{color}
\usepackage{dsfont}
\usepackage{bm}
\usepackage{tikz}
\usepackage{tabularx}
\usepackage{multirow}
\usepackage{booktabs}
\usepackage{xcolor}
\usepackage{subcaption}
\usepackage{afterpage}
\usepackage{svg}
\usepackage{pifont}
\usepackage{dirtytalk}
\usepackage[T1]{fontenc}

\DeclareMathOperator{\cossim}{sim}
\newcommand{\xmark}{\ding{55}}%

\newcommand{\red}[1]{{\color{red}{#1}}}
\newcommand{\blu}[1]{{\color{blue}{#1}}}
\newcommand{\gra}[1]{{\color{gray}{#1}}}

\def\BibTeX{{\rm B\kern-.05em{\sc i\kern-.025em b}\kern-.08em
    T\kern-.1667em\lower.7ex\hbox{E}\kern-.125emX}}
\begin{document}

\makeatletter
\newcommand{\linebreakand}{%
      \end{@IEEEauthorhalign}
      \hfill\mbox{}\par
      \mbox{}\hfill\begin{@IEEEauthorhalign}
}
\makeatother

\newcommand\blfootnote[1]{%
  \begingroup
  \renewcommand\thefootnote{}\footnote{#1}%
  \addtocounter{footnote}{-1}%
  \endgroup
}
\title{Self-supervised Benchmark Lottery on ImageNet: Do Marginal Improvements Translate\\to Improvements on Similar Datasets?}

\author{
\IEEEauthorblockN{Utku Ozbulak$^{1,\,2\,\ddag}$ \quad Esla Timothy Anzaku$^{1,\,2,\,\ddag}$ \quad Solha Kang$^{1}$  \quad Wesley De Neve$^{1,\,2}$ \quad Joris Vankerschaver$^{1,\,3}$}
\IEEEauthorblockA{
\textit{$^1$Center for Biosystems and Biotech Data Science , Ghent University Global Campus, Republic of Korea}\\
\textit{$^2$Department of Electronics and Information Systems, Ghent University, Belgium}\\
\textit{$^3$Department of Applied Mathematics, Computer Science and Statistics, Ghent University, Belgium}\\
\ddag \,These authors contributed equally to this work \\
\{utku.ozbulak,eslatimothy.anzaku,solha.kang,wesley.deneve,joris.vankerschaver\}@ghent.ac.kr}
}

\maketitle




\begin{abstract}
Machine learning (ML) research strongly relies on benchmarks in order to determine the relative effectiveness of newly proposed models. Recently, a number of prominent research effort argued that a number of models that improve the state-of-the-art by a small margin tend to do so by winning what they call a \say{benchmark lottery}. An important benchmark in the field of machine learning and computer vision is the ImageNet where newly proposed models are often showcased based on their performance on this dataset. Given the large number of self-supervised learning (SSL) frameworks that has been proposed in the past couple of years each coming with marginal improvements on the ImageNet dataset, in this work, we evaluate whether those marginal improvements on ImageNet translate to improvements on similar datasets or not. To do so, we investigate twelve popular SSL frameworks on five ImageNet variants and discover that models that seem to perform well on ImageNet may experience significant performance declines on similar datasets. Specifically, state-of-the-art frameworks such as \texttt{DINO} and \texttt{Swav}, which are praised for their performance, exhibit substantial drops in performance while \texttt{MoCo} and \texttt{Barlow Twins} displays comparatively good results. As a result, we argue that otherwise good and desirable properties of models remain hidden when benchmarking is only performed on the ImageNet validation set, making us call for more adequate benchmarking. To avoid the \say{benchmark lottery} on ImageNet and to ensure a fair benchmarking process, we investigate the usage of a unified metric that takes into account the performance of models on other ImageNet variant datasets.

\end{abstract}


\section{Introduction}

\begin{figure}[t!]
\includegraphics[width=0.49\textwidth]{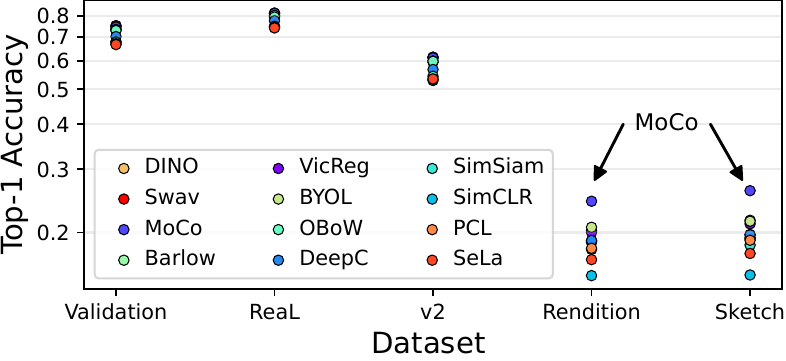}
\includegraphics[width=0.49\textwidth]{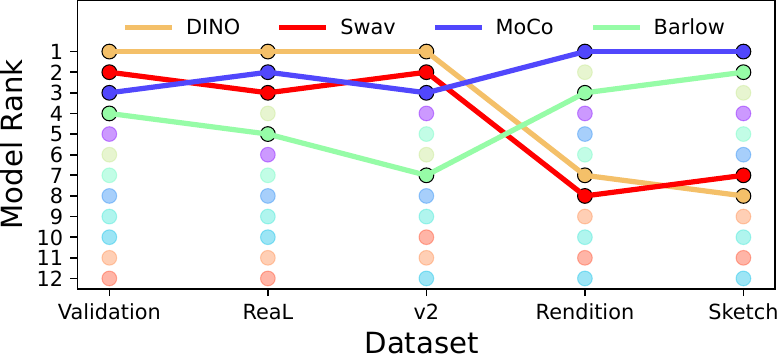}
\caption{(top) Top-1 accuracy of SSL models plotted for variants of ImageNet and (bottom) changes in the top-1 accuracy ranking across the twelve evaluated models, highlighting the performance of \texttt{DINO}, \texttt{Swav}, \texttt{MoCo}, and \texttt{Barlow TWins}. Note that while \texttt{DINO} and \texttt{Swav} occupy top two spots in ImageNet validation, their performance does not translate to Rendition and Sketch datasets in which \texttt{MoCo} and \texttt{Barlow} show tremendous improvements.}
\label{fig:selling_figure}
\end{figure}

Given a large amount of labeled data, deep neural networks (DNNs) were shown to find highly accurate solutions to complex vision problems~\cite{Alexnet,VGG,resnext}. Unfortunately, for many domains, the availability of labeled data is a commodity in short supply due to (1) expert knowledge required to create annotations, (2) the cost associated with labeling, and (3) potential inconsistencies in the labeling process~\cite{labeling_hard_1}. Traditionally, transfer and few-shot learning methods have been the go-to methods when faced with a shortage of labeled data~\cite{wang2020generalizing,transfer_learning}. In recent years, transfer learning in particular has become so popular that studies often employ pretrained models to overcome data shortages. As a result of the straightforward applicability of transfer learning, most deep learning frameworks now come with a large catalogue of models~\cite{pytorch-whitepaper} pretrained on ImageNet~\cite{ILSVRC15:rus}.

A novel methodology that also allows mitigating label shortage is self-supervised learning, where feature extractors are trained in a self-supervised manner on all available data, and without labels~\cite{Simclr,BYOL}. Although this line of work is still in its infancy for supervised computer vision tasks, the results obtained with recent SSL frameworks show promise, achieving comparable results to models trained with supervised learning. On the other hand, while transfer learning is convenient and easy to utilize, self-supervised learning is a slow and expensive process that requires a large amount of compute with backbones requiring a great number of epochs to train, coupled with extensive parameter search requirements for linear probing~\cite{BYOL,Mocov1}. Due to these compute requirements, comparing a collection of SSL frameworks to one another means training multiple models for weeks, if not months, with a typical 8-GPU setup if the dataset under consideration is at the scale of ImageNet. To make matters even more complicated, within the span of two years, more than a hundred unique SSL frameworks were proposed, each claiming state-of-the-art results on ImageNet~\cite{vicreg,barlow,pcl,Swav,Pirl}. For an overview of SSL frameworks and results, see \cite{2023-balestriero-CookbookSelfSupervisedLearning,2023-ozbulak-KnowYourSelfsupervised} and the references therein.

Given the SSL training limitations outlined above, what is the best approach to select a particular SSL framework? The most straightforward answer to this question is to select that SSL framework that achieves the highest top-1 accuracy on ImageNet, since ImageNet is the de facto standard for SSL evaluation in the domain of computer vision. However, most of the recently proposed SSL frameworks only improve the ImageNet validation accuracy marginally (approximately by $1\%$) compared to their predecessor. In this context, a fair question to pose is \say{Are these improvements achieved by chance, or are they thanks to enhancements in the SSL method?} Recent research efforts on the topic of benchmarking and marginal improvements argue that a significant portion of the marginal improvements in DNNs over certain benchmarks arises from improper experimental settings, such as training without a validation set (i.e., using the test set for feedback) and the random initialization of networks~\cite{musgrave2020metric,dehghani2021benchmark}. Unfortunately, the cost to train neural networks, especially in self-supervised settings makes it unfeasible to reproduce majority of the research since training times are measured with years on a single GPU~\cite{Mocov3}.

Given the aforementioned limitations, in this paper, we propose the usage of ImageNet variants along with the ImageNet validation dataset to benchmark SSL models for a fairer and more appropriate evaluation. To do so, we perform large-scale experiments using twelve SSL models on five ImageNet variants. Whereas some of these variant datasets contain similar images to that of ImageNet, others contain images with unique traits, thus enabling a variety of analyses of the generalization power of SSL models. Based on our experiments, we can briefly summarize our novel observations as follows:

\begin{itemize}
\item As summarized in Fig.~\ref{fig:selling_figure}, our experiments indicate that the performance and potential of otherwise good models with desirable properties may remain hidden when evaluations are solely conducted on the ImageNet validation set. Consequently, we assert that benchmarking SSL models across a range of ImageNet variants is essential to understand their true potential.

\item Our experiments reveal that slight improvements in performance on the ImageNet validation do not necessarily result in corresponding improvements on ImageNet variant datasets. In light of this, we advocate for a comprehensive approach that incorporates all five ImageNet variants into the benchmarking process. By doing so, we aim to create a more robust and nuanced evaluation framework that better reflects the diverse conditions posed by these variants.

\item Recognizing the need for a clear and effective metric to measure the performance of neural networks on these diverse datasets, we explore a range of statistical methods with the goal of harmonizing the benchmarks derived from ImageNet variant datasets.
\end{itemize}

\section{Related Work}
\label{sec:SSL}

\subsection{Self-supervised Learning}

The idea of self-supervision dates back to the works of~\cite{becker1992self,de1994learning} where the goal of those works was to achieve an agreement on representations coming from differently transformed views of the same image. After three decades of research efforts, self-supervised training methods are now considered go-to approaches for a large number of use-cases such as image colorization~\cite{image_colorization,image_colorization2}, image super resolution~\cite{super_resolution}, inpainting~\cite{inpainting}, and geometric transformations~\cite{geometric_trans}. Although DNNs trained in self-supervised fashion with such methods were useful in solving these problems, their usage for downstream supervised tasks were shown to be ineffective~\cite{Swav}.

Nevertheless, research efforts on self-supervised training for supervised classification have shown gradual improvements over the course of the last decade~\cite{ssl_early,ssl_early2} and recently, the work of~\cite{deep_cluster} demonstrated that it is possible to train feature extractors on ImageNet in a self-supervised fashion using clustering techniques and achieve comparable results to supervised models. This approach not only inspired future studies which improved SSL training that relies on clustering approaches~\cite{Swav,sela}, it also paved the way for contrastive~\cite{Simclr,Mocov1,Pirl} and distillation-based~\cite{Obow,Simsiam,BYOL,Dino} methods.

Before a DNN can be utilized in a traditional way for downstream predictive tasks, its backbone (i.e., feature extractor) goes through a self-supervised training stage where a neural network backbone is trained in a self-supervised fashion so that it learns \textit{good} representations. We now briefly detail the various SSL frameworks used in this study.

\textbf{Clustering}\,\textendash\, Early works in this space such as \texttt{DeepC}~\cite{deep_cluster} and \texttt{SeLa}~\cite{sela} train backbones with the cross entropy loss (CE). However, instead of using actual labels for the loss calculation, they make use of pseudo-labels that are determined by cluster assignments~\cite{feature_quantization}. Different from \texttt{DeepC} and \texttt{SeLa}, \texttt{Swav} makes use of the following loss
\begin{eqnarray}
\label{eq:swav_swap}
L_{SW} := \ell(\bm{z}_i, \bm{q}_j) + \ell(\bm{z}_j, \bm{q}_i) \,,
\end{eqnarray}
where the aim is to predict the code ($\bm{q}_{\{i,j\}}$) of a view from the representation of another view ($\bm{z}_{\{j,i\}}$) with $\ell(\cdot)$ measuring the fit between the two.

\begin{table}[t]
\centering
\caption{Datasets used in this study and their properties.}
\scriptsize
\begin{tabular}{lcccc}
\cmidrule[1pt]{1-5}
Dataset & \phantom{-}Image count\phantom{-} & \phantom{-}Classes\phantom{-} & \phantom{-}Image per class\phantom{-} & \phantom{-}Multi-label\phantom{-} \\
\cmidrule[0.5pt]{1-5}
{Validation} & 50,000 & 1,000 & 50 & \xmark \\
{ReaL} & 50,000 & 1,000 & 50 & \checkmark  \\
{v2} & 10,000 & 1,000 & 10 & \xmark  \\
{Rendition} & 30,000 & 200 & $\sim$150 & \xmark  \\
{Sketch} & 50,889 & 1,000 & $\sim$50 & \xmark \\
{Adversarial} & 7,500 & 200 & $\sim$37 & \xmark \\
\cmidrule[1pt]{1-5}
\end{tabular}
\label{tbl:imagenet_dataset}
\end{table}

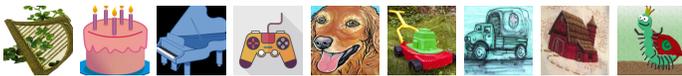
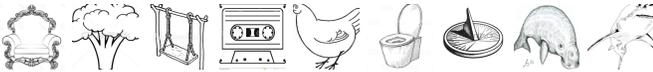
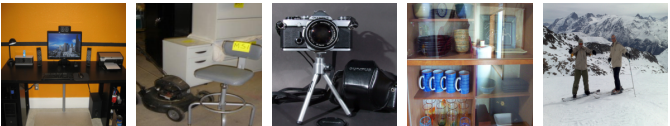
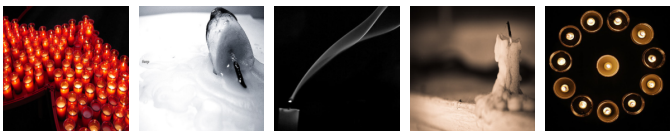
\begin{figure}[t!]
\begin{subfigure}{0.5\textwidth}
\centering
    \includegraphics[width=.1\textwidth]{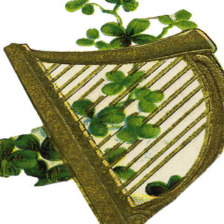}\hfill
    \includegraphics[width=.1\textwidth]{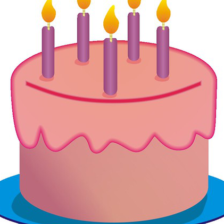}\hfill
    \includegraphics[width=.1\textwidth]{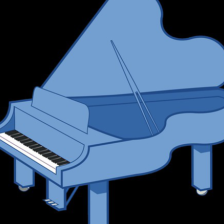}\hfill
    \includegraphics[width=.1\textwidth]{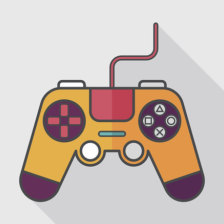}\hfill
    \includegraphics[width=.1\textwidth]{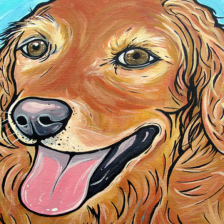}\hfill
    \includegraphics[width=.1\textwidth]{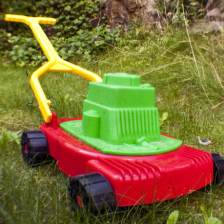}\hfill
    \includegraphics[width=.1\textwidth]{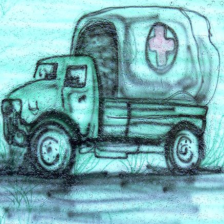}\hfill
    \includegraphics[width=.1\textwidth]{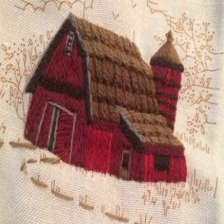}\hfill
    \includegraphics[width=.1\textwidth]{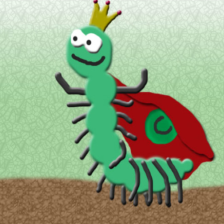}
    \caption{Sample images from ImageNet Rendition}
    \label{fig:imagenet_rendition_examples}
\vspace{0.5em}
\end{subfigure}
\begin{subfigure}{0.48\textwidth}
    \includegraphics[width=.1\textwidth]{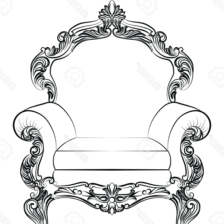}\hfill
    \includegraphics[width=.1\textwidth]{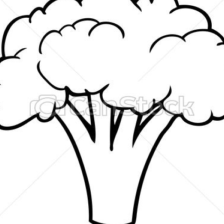}\hfill
    \includegraphics[width=.1\textwidth]{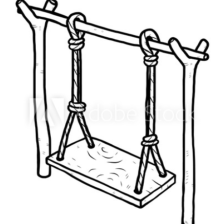}\hfill
    \includegraphics[width=.1\textwidth]{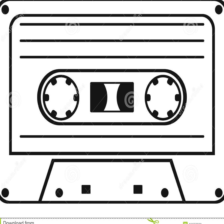}\hfill
    \includegraphics[width=.1\textwidth]{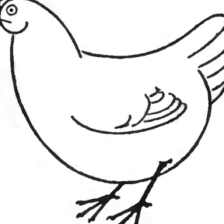}\hfill
    \includegraphics[width=.1\textwidth]{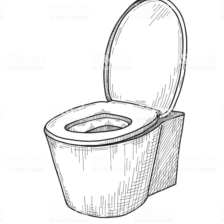}\hfill
    \includegraphics[width=.1\textwidth]{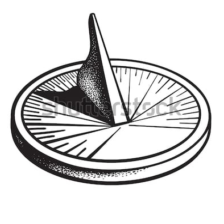}\hfill
    \includegraphics[width=.1\textwidth]{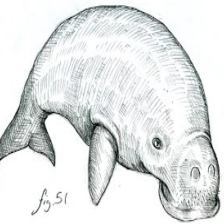}\hfill
    \includegraphics[width=.1\textwidth]{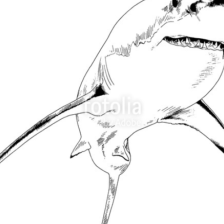}
    \caption{Sample images from ImageNet Sketch}
    \label{fig:imagenet_sketch_examples}
\vspace{0.5em}
\end{subfigure}

\begin{subfigure}{0.48\textwidth}
\begin{tikzpicture}
\def\sety1{0}
\node[inner sep=0pt] () at (0,\sety1)
{\includegraphics[width=.19\textwidth]{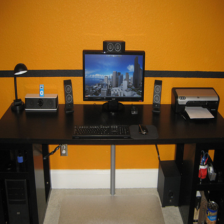}};
\node[inner sep=0pt] () at (1.8,\sety1)
{\includegraphics[width=.19\textwidth]{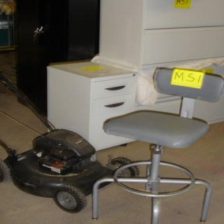}};
\node[inner sep=0pt] () at (3.6,\sety1)
{\includegraphics[width=.19\textwidth]{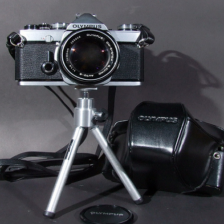}};
\node[inner sep=0pt] () at (5.4,\sety1)
{\includegraphics[width=.19\textwidth]{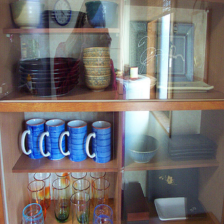}};
\node[inner sep=0pt] () at (7.2,\sety1)
{\includegraphics[width=.19\textwidth]{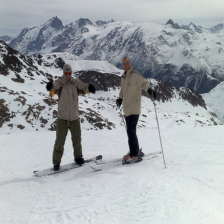}};

\node [] at (0,-1) () {\scriptsize {Monitor}, {Desk}};
\node [] at (0,-1.3) () {\scriptsize {Mouse}, {Printer}};

\node [] at (1.8,-1) () {\scriptsize {File cabinet}};
\node [] at (1.8,-1.3) () {\scriptsize {Mower}, {Vacuum}};

\node [] at (3.6,-1) () {\scriptsize {Lens cover}, {Tripod}};
\node [] at (3.6,-1.3) () {\scriptsize {Reflex camera}};

\node [] at (5.4,-1) () {\scriptsize {Cabinet}, {Cup}};
\node [] at (5.4,-1.3) () {\scriptsize {Coffee mug}};

\node [] at (7.2,-1) () {\scriptsize {Alp}, {Ski}};
\end{tikzpicture}
\caption{Sample images from ImageNet validation and their ReaL labels.}
\label{fig:imagenet_real_examples}
\vspace{0.5em}
\end{subfigure}

\begin{subfigure}{0.48\textwidth}
\begin{tikzpicture}
\def\sety1{0}
\node[inner sep=0pt] () at (0,\sety1)
{\includegraphics[width=.19\textwidth]{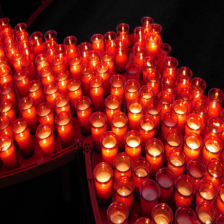}};
\node[inner sep=0pt] () at (1.8,\sety1)
{\includegraphics[width=.19\textwidth]{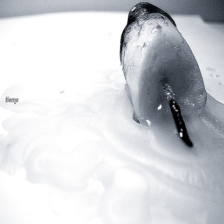}};
\node[inner sep=0pt] () at (3.6,\sety1)
{\includegraphics[width=.19\textwidth]{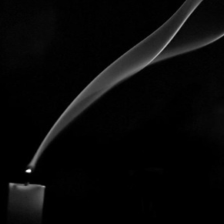}};
\node[inner sep=0pt] () at (5.4,\sety1)
{\includegraphics[width=.19\textwidth]{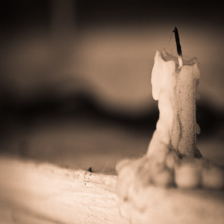}};
\node[inner sep=0pt] () at (7.2,\sety1)
{\includegraphics[width=.19\textwidth]{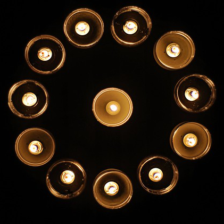}};

\end{tikzpicture}
\caption{Sample images from ImageNet Adversarial where these images are misclassified by a ResNet-50 into labels other than \textit{candle}.}
\label{fig:imagenet_adv_example_1}
\end{subfigure}

\caption{Example images from ImageNet variants.}
\end{figure}

\textbf{Contrastive}\,\textendash\,Moving towards the contrastive-based methods, \texttt{SimCLR}~\cite{Simclr} and \texttt{MoCo}~\cite{Mocov3} both make use of InfoNCE loss where this loss tries to pull together representations of positive pairs, while contrasting negative pairs~\cite{infonce_loss}. Given a batch of $n$ images $[\bm{x}_1, \ldots, \bm{x}_n]$, \texttt{SimCLR} generates an augmented versions of each image, thus resulting with $2n$ images $\mathtt{B} = [\bm{x}_1, \bm{x}_1^{(a)}, \ldots, \bm{x}_n, \bm{x}_n^{(a)}]$  and minimizes the InfoNCE loss for each positive pair as follows:
\begin{eqnarray}
\label{eq:info}
L_{CON} := -\log \frac{\exp(\cossim(\mathtt{B}_i, \mathtt{B}_j)/ \tau)}{\sum_{k=1}^{2n} \mathds{1}_{\{k\neq i\}} \exp(\cossim(\mathtt{B}_i, \mathtt{B}_k)/ \tau)} \,.
\end{eqnarray}
Above, $\mathds{1}_{\{k\neq i\}}$ denotes the indicator function which is used to select negative pairs and $\cossim(\cdot, \cdot)$ denotes the cosine similarity between two features. Building upon the InfoNCE loss, \texttt{PCL}~\cite{pcl} uses a novel loss called ProtoNCE that is derived from InfoNCE and trains backbones with it.

\textbf{Distillation}\,\textendash\ SSL training that rely on distillation use a variety of well-known error estimates to enable training. \texttt{BYOL}~\cite{BYOL} employs mean squared error between the normalized predictions and target projections while \texttt{SimSiam}~\cite{Simsiam} simply uses cosine similarity to train backbones. \texttt{DINO} integrates CE with local and global crops of images and \texttt{OBoW}~\cite{Obow} extends the SSL framework proposed by~\cite{gidaris2020learning} which uses bag-of-visual-words.

\textbf{Information-maximization}\,\textendash\ Novelty of frameworks under this category entirely rely on their unique and innovative loss functions. For example, \texttt{Barlow}~\cite{barlow} uses the following loss:
\begin{eqnarray}
\label{eq:loss_barlow}
L_{BT} := \underbrace{\sum_i (1 - \mathcal{C}_{ii})^{2}}_\text{Invariance} +  \underbrace{\lambda \sum_i \sum_{i\neq j} \mathcal{C}_{ij}^2}_\text{Redundancy reduction} \,,
\end{eqnarray}
where $\mathcal{C}$ denotes a cross-correlation matrix. \texttt{VicReg} further expands the above loss into three different terms: (1) variance (2) invariance, and (3) covariance~\cite{vicreg}.

\subsection{Benchmarking in Computer Vision}

Benchmarking plays a pivotal role in the field of machine learning, serving as a cornerstone for evaluating the performance and efficiency of various algorithms, models, and systems. In the dynamic landscape of artificial intelligence, where novel approaches and architectures are continually emerging, benchmarking provides a standardized framework to assess and compare their effectiveness. Numerous benchmarking initiatives have been instrumental in shaping the discourse surrounding machine learning performance. Prominent examples include datasets such as MNIST~\cite{lecun1998gradient}, ImageNet~\cite{ILSVRC15:rus}, and COCO~\cite{coco}, which have become standard benchmarks various tasks.

As the number of datasets, architectures, and unique SSL frameworks grew, the number of studies that perform a comparative analysis of each also increased~\cite{zhai2019large,goyal2019scaling,kornblith2019better,how_well_transfer}. The closest work to ours in the literature is ~\cite{how_well_transfer} where the authors investigate the generalization power of contrastive- and clustering-based SSL methods on downstream tasks (with re-training on those tasks). Contrary to their work, we do not retrain the models on an additional dataset and investigate the generalization power of ImageNet-trained SSL models on a variety of datasets created with the inspiration taken from ImageNet~\cite{imagenet_a_o,imagenet_sketch,imagenet_r,imagenet_r_evaluation,imagenetv2}, where our work borders the field of domain adaptation~\cite{zhang2020impact,zhang2021semi} and out-of-distribution generalization~\cite{hendrycks2016baseline,imagenet_r,winkens2020contrastive}.

While benchmarking serves as a crucial tool for evaluating and comparing models, it is not without its challenges. ImageNet has been more robust to some of these challenges regarding benchmarking, thanks to its sheer size; however, it is not without problems. One issue with ImageNet is its bias towards certain classes, such as those under the \say{canine} (i.e., dogs) label, which account for 130 out of 1,000 classes~\cite{ozbulak2021evaluating}. Another issue with ImageNet is the unavailability of the test set, meaning that all benchmarking on ImageNet is performed on the validation dataset during training. This experimental procedure may lead to cases known as \say{overfitting to the benchmark}: situations in which a model is optimized specifically for a particular benchmark or dataset, to the extent that its performance becomes overly specialized and might not generalize well to new or unseen data. In this work, we tackle this issue and analyze the performance of popular SSL models on various ImageNet-like datasets to identify whether their performance on ImageNet validation translates to those datasets or not.

\section{Datasets and Models}
\subsection{Datasets}
\label{sec:Datasets}

In Table~\ref{tbl:imagenet_dataset}, we provide an overview of the datasets used to measure the properties of ImageNet-trained SSL models in this study, detailing the image count, number of classes as well as image per class. In what follows, we provide brief description of these datasets.

\textbf{ImageNet}\,\textendash\,One of the most-used datasets to evaluate SSL training, ImageNet~\cite{ILSVRC15:rus} has played a crucial role in measuring progress in the field of computer vision~\cite{Alexnet,VGG,resnet,Inceptionv1}. The ImageNet training set contains more than a million images spread across $1,000$ classes and the validation set contains $50$ images per class, totalling up to $50,000$ images.

\textbf{ImageNet ReaL}\,\textendash\,To the dismay of the computer vision community, a number of studies indicate that ImageNet validation set has some issues with the labels, whereby a portion of the images suffer from having a single label per image, restrictive label proposals, and arbitrary class distinctions~\cite{imagenet_problem1,imagenet_problem2}. To overcome this problem,~\cite{ImagenetReal} came up with a new, more robust labeling strategy that alleviates the aforementioned issues and named the new set of labels as {ReaL}, where this new set of labels can contain up to $10$ labels for a single image.

In Figure~\ref{fig:imagenet_real_examples} we provide example ReaL labels instead of the original ImageNet validation labels. Note that {ReaL} labels converts the ImageNet validation set from a single-label dataset to a multi-label dataset where having a prediction as any of the given {ReaL} classes is considered a correct one. Furthermore, {ReaL} also removes a number of images from the ImageNet validation for being ambiguous and not falling into any of the categories employed in ImageNet-1k.

\textbf{ImageNet v2}\,\textendash\,In a quest to evaluate the correctness of using ImageNet validation as a single source of model performance,~\cite{imagenetv2} replicated the image collection methodology of the ImageNet and assembled a new validation set called {ImageNet v2}. This dataset comes with $10$ images per class and contains $10,000$ images in total and can be seen as in-distribution dataset that similar to the ImageNet validation.

\textbf{ImageNet Rendition}\,\textendash\,In order to understand the out-of-distribution generalization of DNNs trained on ImageNet and to analyze the impact of real-world distribution shifts such as changes in image style, blurriness, geographic location, and camera view,~\cite{imagenet_r} collected {ImageNet Rendition} dataset which contains $30,000$ images for a subset of $200$ classes of ImageNet (see Fig~\ref{fig:imagenet_rendition_examples} for a set of example images).

\textbf{ImageNet Sketch}\,\textendash\,Local receptive fields in DNNs are known to be extremely influential when making predictions to the point where DNNs sometimes ignore global features such as the shape~\cite{dnn_texture_bias}. In order to evaluate the local biases of DNNs trained on ImageNet, ~\cite{imagenet_sketch} released a sketch variant of ImageNet that is void of color and texture. This dataset contains approximately $50$ images per class, totalling up to $50,889$ images (see Fig~\ref{fig:imagenet_sketch_examples} for a set of example images).

\textbf{ImageNet Adversarial}\,\textendash\,A recently discovered flaw of DNNs is their vulnerability to adversarial examples~\cite{Goodfellow-expharnessing,IFGS} where carefully performed small perturbations on inputs cause large changes in the predictions. Inspired by this phenomenon,~\cite{imagenet_a_o} curated a dataset of $7,500$ images that contains naturally occurring adversarial images (i.e., images that contain mild visual tricks) that are misclassified by most DNNs. In Fig.~\ref{fig:imagenet_adv_example_1} we provide example natural adversarial examples for the ImageNet class \say{candle}.

\subsection{Models}

The SSL frameworks described in Section~\ref{sec:SSL} often come with repositories which contain feature extractors pretrained on the ImageNet training set. For a faithful analysis, we take the parameters of ResNet-50 backbone networks as-is, without making any changes from the respective repositories of the SSL frameworks.

Unfortunately, the aforementioned SSL backbones often do not contain linear layers. In order to enable straightforward inference, we undertake the task of individually training each of these frameworks on the ImageNet training set. During this training process, we adhere to the SGD training routines outlined in the respective papers for each SSL framework. For additional information on this subject, we direct the reader to the section that elaborates on \say{Linear Evaluation} in the corresponding papers of the SSL frameworks.

\begin{table}[t!]
\scriptsize
\centering
\caption{Top-1 accuracies (in percentage) of various self-supervised models when evaluated on ImageNet validation as well as different ImageNet variant datasets.}
\begin{tabular}{lc|ccccc}
\toprule
Model & Val. & ReaL & v2 & Rendition & Sketch & Adv. \\
 \cmidrule[0.25pt]{1-7}
\texttt{DINO}  & 75.1 & 81.4 & 61.4 & 18.9 & 19.2 & 2.3 \\
\texttt{Swav}  & 75.0 & 81.3 & 61.3 & 18.7 & 19.3 & 2.4 \\
\texttt{MoCo}  & 74.4 & 81.3 & 61.1 & 24.4 & 26.1 & 1.8 \\
\texttt{Barlow} & 73.2 & 80.0 & 59.6 & 20.2 & 21.6 & 1.5 \\
\texttt{VicReg} & 73.2 & 79.7 & 59.8 & 20.0 & 21.1 & 1.6 \\
\texttt{BYOL} & 73.1 & 80.1 & 59.7 & 20.6 & 21.4 & 1.6 \\
\texttt{OBoW} & 72.8 &  79.4 & 59.8 & 18.9 & 19.7 & 3.3  \\
\texttt{DeepC} & 70.0 & 77.5 & 56.7 & 19.0 & 19.6 & 1.4 \\
\texttt{SimSiam} & 67.8 & 74.8 & 54.3 & 17.9 & 18.4 & 1.2\\
\texttt{SimCLR} & 67.5 & 74.7 & 52.8 & 15.1 & 15.2 & 1.1 \\
\texttt{PCL} & 66.8 & 74.2 & 53.3 & 18.0 & 19.0 & 1.3 \\
\texttt{SeLa} & 66.5 & 74.0 & 53.4 & 16.8 & 17.4 & 1.2\\
\bottomrule
\end{tabular}
\label{tbl:acc_table}
\end{table}

\begin{table}[t!]
\scriptsize
\centering
\caption{The performance of self-supervised models is ranked based on their initial positions on the ImageNet validation set. Similarly, for ImageNet variant datasets, we rank the models based on their performance (as depicted in Table~\ref{tbl:acc_table}) and highlight changes in blue (indicating a better rank compared to ImageNet validation) and red (indicating a worse rank compared to ImageNet validation).}
\begin{tabular}{lc|ccccc}
\toprule
Model & Val. & ReaL & v2 & Rendition & Sketch & Adv. \\
 \cmidrule[0.25pt]{1-7}
\texttt{DINO} & 1  &  1 (\texttt{\gra{+0}}) & 1 (\texttt{\gra{+0}}) & 7 (\texttt{\red{-6}}) & 8 (\texttt{\red{-7}}) & 3 (\texttt{\red{-2}})  \\
\texttt{Swav} & 2  &  3 (\texttt{\red{-1}}) & 2 (\texttt{\gra{+0}}) & 8 (\texttt{\red{-6}}) & 7 (\texttt{\red{-5}})  & 2 (\texttt{\gra{+0}}) \\
\texttt{MoCo} & 3 &  2 (\texttt{\blu{+1}}) & 3 (\texttt{\gra{+0}}) & 1 (\texttt{\blu{+2}}) & 1 (\texttt{\blu{+2}}) & 4 (\texttt{\red{-1}}) \\
\texttt{Barlow} & 4 & 5 (\texttt{\red{-1}}) & 7 (\texttt{\red{-3}})  & 3 (\texttt{\blu{+1}}) & 2 (\texttt{\blu{+2}}) & 7 (\texttt{\red{-3}}) \\
\texttt{VicReg} & 5 &  6 (\texttt{\red{-1}}) & 4 (\texttt{\blu{+1}})  & 4 (\texttt{\blu{+1}}) & 4 (\texttt{\blu{+1}}) & 5 (\texttt{\gra{+0}})  \\
\texttt{BYOL} & 6 &   4 (\texttt{\blu{+2}}) & 6 (\texttt{\gra{+0}}) & 2 (\texttt{\blu{+4}}) & 3 (\texttt{\blu{+3}}) & 6 (\texttt{\gra{+0}}) \\
\texttt{OBoW} & 7 &  7 (\texttt{\gra{+0}}) & 5 (\texttt{\blu{+2}}) & 6 (\texttt{\blu{+1}}) & 5 (\texttt{\blu{+2}}) & 1 (\texttt{\blu{+6}}) \\
\texttt{DeepC} & 8 & 8 (\texttt{\gra{+0}}) & 8 (\texttt{\gra{+0}}) & 5 (\texttt{\blu{+3}}) & 6 (\texttt{\blu{+2}})  & 8 (\texttt{\gra{+0}}) \\
\texttt{SimSiam} & 9 & 9 (\texttt{\gra{+0}}) & 9 (\texttt{\gra{+0}}) & 10 (\texttt{\red{-1}}) & 10 (\texttt{\red{-1}}) & 11 (\texttt{\red{-2}}) \\
\texttt{SimCLR} & 10 & 10 (\texttt{\gra{+0}}) & 12 (\texttt{\red{-2}}) & 12 (\texttt{\red{-2}}) & 12 (\texttt{\red{-2}}) & 12 (\texttt{\red{-2}}) \\
\texttt{PCL} & 11 & 11 (\texttt{\gra{+0}}) & 11 (\texttt{\gra{+0}}) & 9 (\texttt{\blu{+2}}) & 9 (\texttt{\blu{+2}}) & 9 (\texttt{\blu{+2}}) \\
\texttt{SeLa} & 12 & 12 (\texttt{\gra{+0}}) & 10 (\texttt{\blu{+2}}) & 11 (\texttt{\blu{+1}}) & 11 (\texttt{\blu{+1}}) & 10 (\texttt{\blu{+2}}) \\
\bottomrule
\end{tabular}
\label{tbl:rank_table}
\end{table}

\section{Experimental Results}

In Table~\ref{tbl:acc_table}, we present the top-1 accuracy of each SSL backbone on the ImageNet validation set (refer to the second column). We have sorted the models based on their performance on the ImageNet validation set and maintain this order throughout the paper for consistency. For the majority of the models, we successfully capture the reported top-1 accuracy in their respective benchmarks, and for those where we do not, the models we obtain have only minor discrepancies.

In Table~\ref{tbl:acc_table}, we also include the top-1 accuracy performance of models on five ImageNet variant datasets. For ease of comparing model performances relative to each other, we introduce Table~\ref{tbl:rank_table}. In this table, we rank the models based on their initial positions on the ImageNet validation set as well as the ImageNet variants. To enhance clarity, we use a blue font to highlight better rankings (compared to the initial standing on ImageNet validation) and a red font to indicate worse rankings. 

We will now discuss the implications of the results obtained for each variant dataset.

\subsection{Performance on ImageNet Variants}

\textbf{ImageNet ReaL}\,\textendash\,When evaluating the models using ReaL labels, the newly acquired results are presented in Table~\ref{tbl:acc_table}. In contrast to certain supervised models assessed in the study by~\cite{ImagenetReal}, we did not observe a decline in accuracy for any of the SSL models. This emphasizes the robustness of SSL models, especially for those with lower initial accuracy, when subjected to evaluations with ReaL labels. Additionally, we note that the linear accuracy gains for SSL models tend to be more pronounced for less accurate models compared to their more accurate counterparts. Consequently, when assessed with ReaL labels, the disparity between the best- and worst-performing SSL models decreases from $8.7\%$ (on the ImageNet validation set) to $7.4\%$ (on ImageNet ReaL).  

\textbf{ImageNet v2}\,\textendash\,In the work of ~\cite{imagenetv2}, supervised models are reported to have a reduction in top-1 accuracy within the range of $11\%$ to $14\%$ when evaluated on ImageNet-v2. Likewise in Table~\ref{tbl:acc_table}, we observe that SSL models also show identical drops in accuracy. Similar to the evaluation with ReaL labels, we observe a few rank-order differences in v2, indicating once again, that marginal improvements on {ImageNet validation} may be misleading when interpreted as overall improvements on model performance.

\textbf{ImageNet Rendition}\,\textendash\,When the evaluation is performed on the rendition images, we observe significant declines in accuracy across all models, accompanied by substantial changes in their rank-ordering. Notably, the two initially top-performing models, \texttt{DINO} and \texttt{Swav}, are replaced by \texttt{MoCo} and \texttt{BYOL}, with subsequent positions taken by \texttt{Barlow} and \texttt{VicReg}. Particularly noteworthy is the performance of \texttt{MoCo}, which stands out as clearly better than any other model, with the closest competitor showing approximately $4\%$ less top-1 accuracy. Furthermore, the drop in the performance of \texttt{SimCLR} is so pronounced that the difference between the best- and worst-performing models for ImageNet Rendition is larger than that observed in ImageNet validation. This observed shift in rankings highlights the dynamic nature of model performance on various out-of-distribution datasets, emphasizing the importance of understanding the nuances in model behavior and benchmarking.

\textbf{ImageNet Sketch}\,\textendash\,Similar to the results obtained with Rendition, our findings indicate that improvements on ImageNet validation do not seamlessly translate into improvements on the Sketch dataset. Specifically, \texttt{DINO} and \texttt{Swav} exhibit larger declines in accuracy compared to other models, highlighting a consistent trend. Once again, \texttt{MoCo} secures the top position, boasting a remarkable $4.5\%$ top-1 linear accuracy difference compared to the next-most-accurate SSL model. This reinforces \texttt{MoCo}'s resilience and effectiveness across different datasets.

\textbf{ImageNet Adversarial}\,\textendash\,As it can be seen, natural adversarial adversarial examples appear to reduce the accuracy of almost all models down to single digits no matter the training routine. This consistent reduction in accuracy highlights the robustness challenges that exist across different models, suggesting that the vulnerability to (natural) adversarial examples is a pervasive issue that transcends specific SSL training routines.

\begin{figure}[t!]
\centering
\begin{subfigure}{0.33\textwidth}
\includegraphics[width=\textwidth]{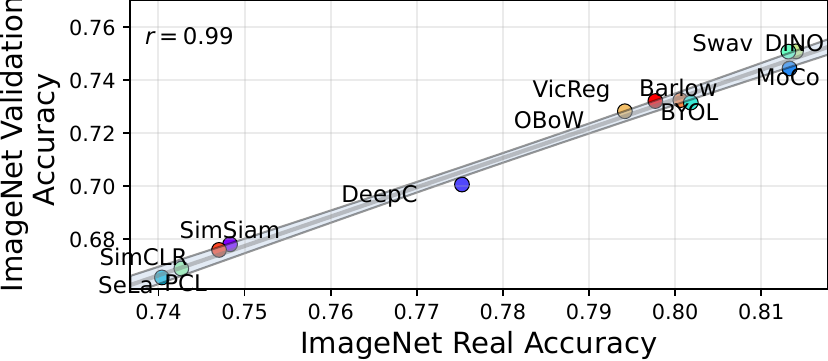}
\label{fig:acc_comp_real}
\end{subfigure}
\begin{subfigure}{0.33\textwidth}
\includegraphics[width=\textwidth]{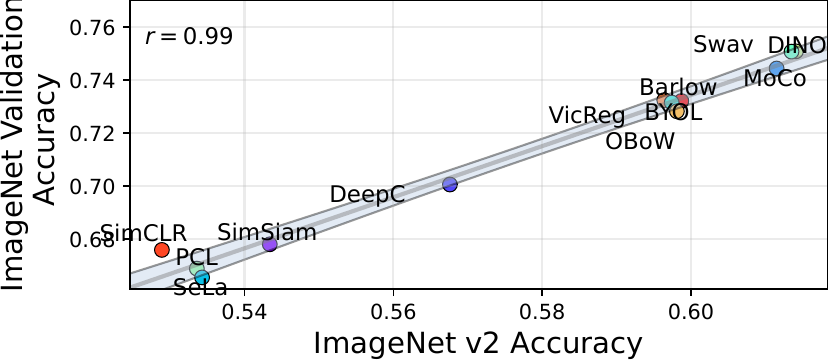}
\label{fig:acc_comp_v2}
\end{subfigure}
\begin{subfigure}{0.33\textwidth}
\includegraphics[width=\textwidth]{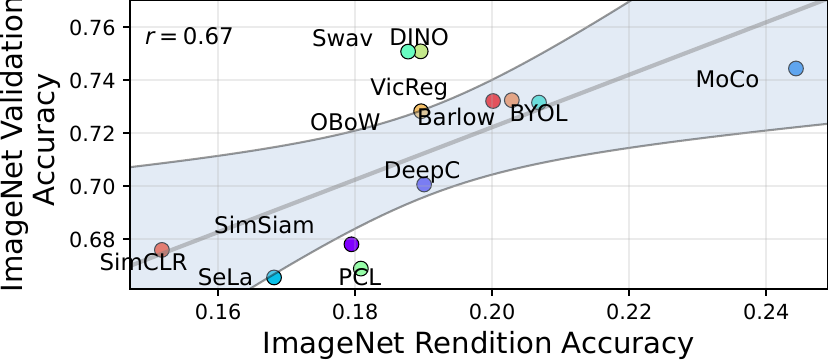}
\label{fig:acc_comp_r}
\end{subfigure}
\begin{subfigure}{0.33\textwidth}
\includegraphics[width=\textwidth]{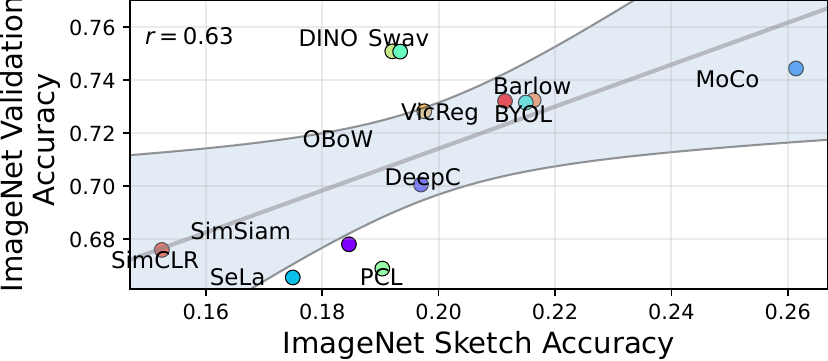}
\label{fig:acc_comp_s}
\end{subfigure}
\begin{subfigure}{0.33\textwidth}
\includegraphics[width=\textwidth]{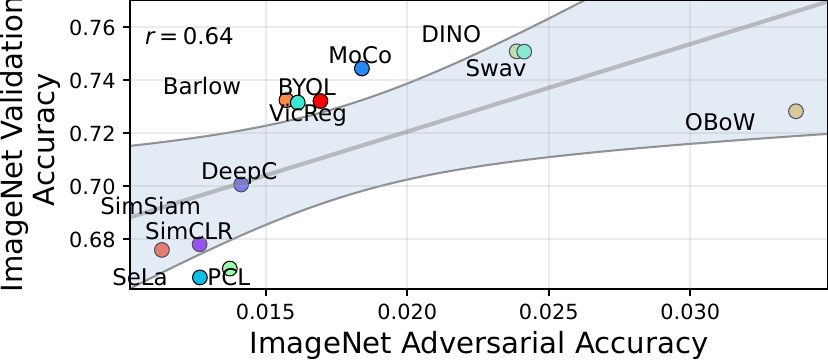}
\label{fig:acc_comp_a}
\end{subfigure}
\caption{Comparison of top-1 linear accuracy on ImageNet Validation to five ImageNet variants. Gray lines in each figure indicate the regression lines while the shaded area depicts the confidence interval. Pearson's correlation coefficient is given on the top-left corner of each figure.}
\label{fig:acc_comp_figs}
\end{figure}

\subsection{Correlation of Performance}

In Figure~\ref{fig:acc_comp_figs}, we present a comparison of the top-1 accuracy of models for various ImageNet variants plotted against the top-1 accuracy obtained on the standard {ImageNet validation} set. For ImageNet-Real and v2, we observe a robust correlation ($r=0.99$) between ImageNet validation set with only minor differences in rank orders, as indicated in Table~\ref{tbl:acc_table}. This high correlation implies that models performing well on ImageNet validation tend to exhibit similar success on the ReaL and v2 datasets.

Conversely, when examining Rendition, Sketch, and Adversarial the correlation notably diminishes to approximately $r=0.6$. This substantial drop suggests that the accuracy achieved on ImageNet validation is not a reliable indicator of a model's performance in out-of-distribution settings. The lower correlation highlights the unique challenges posed by these variant datasets, emphasizing the importance of evaluating model performance in diverse scenarios to ensure robust generalization across different data distributions.

\begin{table}[t!]
\scriptsize
\centering
\caption{Aggregate measures of accuracy across all ImageNet variants. We rank the models based on their aggregate accuracy, with changes compared to the model's rank on ImageNet validation indicated between parentheses.}
\begin{tabular}{lcccc}
\toprule
Model & Weighted Avg. & Rank & Geo. mean & Rank \\
 \cmidrule[0.25pt]{1-5}
\texttt{DINO}  & 42.2 & 5 (\texttt{\red{-4}}) & 21.0 & 4 (\texttt{\red{-3}}) \\
\texttt{Swav}  & 42.1 & 6 (\texttt{\red{-4}}) & 21.2 & 3 (\texttt{\red{-1}}) \\
\texttt{MoCo}  &  45.6 & 1 (\texttt{\blu{+2}}) & 22.4 & 2 (\texttt{\blu{+1}}) \\
\texttt{Barlow} & 42.6 & 3 (\texttt{\blu{+1}}) & 19.8 & 7 (\texttt{\red{-3}}) \\
\texttt{VicReg} & 42.3 & 4 (\texttt{\blu{+1}}) & 20.0 & 6 (\texttt{\red{-1}}) \\
\texttt{BYOL} & 42.7 & 2 (\texttt{\blu{+4}}) & 20.2 & 5 (\texttt{\blu{+1}}) \\
\texttt{OBoW} & 41.6 & 7 (\texttt{\gra{+0}}) & 22.5 & 1 (\texttt{\blu{+6}}) \\
\texttt{DeepC} & 40.6 & 8 (\texttt{\gra{+0}}) & 18.7 & 8 (\texttt{\gra{+0}}) \\
\texttt{SimSiam} & 38.9 & 9 (\texttt{\gra{+0}}) & 17.4 & 10 (\texttt{\red{-1}}) \\
\texttt{SimCLR} & 37.1 & 12 (\texttt{\red{-2}})& 15.8 & 12 (\texttt{\red{-2}}) \\
\texttt{PCL} &  38.9 & 10 (\texttt{\blu{+1}}) & 17.7 & 9 (\texttt{\blu{+2}}) \\
\texttt{SeLa} & 38.0 & 11 (\texttt{\blu{+1}}) & 16.9 & 11 (\texttt{\blu{+1}}) \\
\bottomrule
\end{tabular}
\vspace{-1.5em}
\label{tbl:weighted_geo_avg_table}
\end{table}

\begin{figure}[t!]
\includegraphics[width=.5\textwidth]{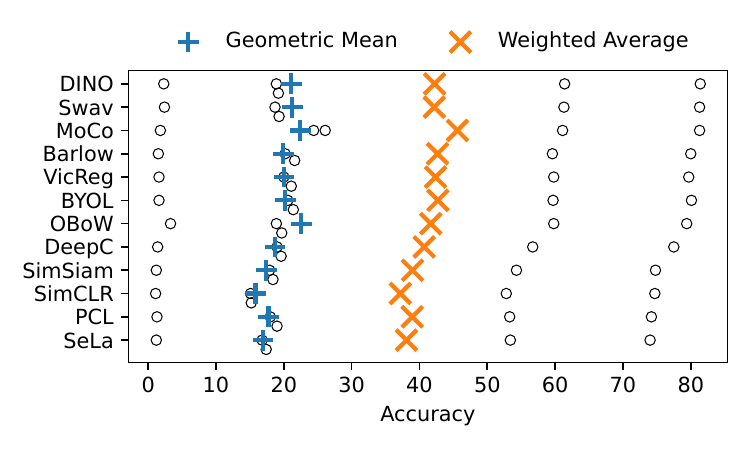}
\caption{Aggregate measures of accuracy for each model across all ImageNet variants. The points in the background represent the accuracy on an individual dataset.}
\label{fig:aggregate_accuracy}
\end{figure}

\subsection{Aggregate Measures of Accuracy}

Building upon the insights gained from the earlier observations, we now investigate aggregate measures of accuracy in order to conduct a comprehensive performance assessment on ImageNet variants. To do so, we employ two distinct aggregate measures of accuracy, namely the weighted average and the geometric mean.

For a given model, the weighted average, $W$, is the average performance across all ImageNet variants, where the accuracy on each dataset is weighted by the number of images in that dataset:
\[
    W = \frac{\sum_{i=1}^k N_i A_i}{\sum_{i=1}^k N_i}, 
\]
with $N_i$ the number of images and $A_i$ the accuracy on the $i$th ImageNet variant. This metric takes into account performance on all datasets, but due to the weight factor poor performance on small datasets is not penalized as harshly as on large datasets.

The choice of the number of images as a weighing factor for the weighted average is reminiscent of the construction of the fixed-effect model in meta-analysis~\cite{2009-borenstein-IntroductionMetaAnalysis}, where the precision (i.e., the inverse of the variance) is used as a weighing factor. In fact, as the precision is proportional to the number of points in the dataset, the fixed-effect estimator for the aggregate effect agrees with our weighted average up to a constant multiplicative factor. This unlocks certain attractive properties that the weighted average is conjectured to have, such as the fact that its precision will be proportional to the sum of the precisions on the individual datasets. To make this link more robust would require quantifying the uncertainty in the accuracy on individual datasets, which we do not attempt in this work. 

The second metric that we consider is the geometric mean, $G$, defined as
\[
    G = \sqrt[k]{A_1 \times \cdots \times A_k},
\]
with $A_i$ again the accuracy on the $i$th ImageNet variant. This is a ``pessimistic'' metric in the sense that small changes in accuracy on low-scoring datasets are more impactful than changes in accuracy of the same magnitude on high-scoring datasets. For example, consider a model with an accuracy of $A_1 = 0.2$ and $A_2 = 0.8$ on two datasets, so that the geometric mean is $G = 0.4$. A decrease in accuracy of $0.1$ units on the first dataset will decrease the geometric mean to $G' = 0.28$, a $30\%$ decrease, while the same decrease on the second dataset results in a geometric mean of $G'' = 0.37$, only a $7.5\%$ decrease.

While the two provided metrics penalize changes in accuracy differently, they nevertheless convey a similar message (see Table~\ref{tbl:weighted_geo_avg_table} and Figure~\ref{fig:aggregate_accuracy}), with \texttt{MoCo} taking the first (weighted average) or second (geometric mean) spot, while \texttt{DINO} and \texttt{Swav} see a drop in aggregate accuracy and hence in rank, compared to their performance on ImageNet validation. The model that scores best under the geometric mean is \texttt{OBoW}, due to its improved performance on ImageNet Adversarial, the dataset on which all models score poorly. This performance is not characteristic of the model as a whole, a fact that is reflected in its mediocre rank under the geometric mean.

\section{Discussion and Conclusion}

Ever since DNNs became mainstream solutions for computer vision problems, the selection of an appropriate model for the task at hand has become one of the important decisions to make. With the introduction of SSL, this problem becomes two-fold: (1) selection of the model and (2) use of the most appropriate SSL framework to train that model. Unfortunately, there are no shortcuts for both decisions. While the ImageNet validation accuracy is indeed helpful in making a decision, as we have shown, making a choice purely based on this criterion can miss certain useful aspects of SSL frameworks. However, this does not mean that the usage of any other ImageNet variant is more appropriate compared to the ImageNet validation set. On the contrary, we observe that each SSL framework comes with its own unique traits. In particular, we have identified:

\begin{itemize}
    \item Failure of \texttt{DINO} and \texttt{Swav} in having their performance translate from ImageNet to its out-of-distribution variants.

    \item Absence of improved performance for \texttt{VicReg} over \texttt{Barlow}, even though the former has been proposed as an enhancement over the latter.

    \item Superb results obtained with \texttt{MoCo} over all other SSL models under dataset shifts.

    \item Improvements in ranking for \texttt{BYOL} and \texttt{Barlow}  under dataset shifts relative to their initial standings.
    
    \item Noticeable resistance of \texttt{OBoW} against adversarial exampled compared to other models.
    
    \item No correlation between performance (measured with the accuracy across ImageNet variants) and the core self-supervised training methodology: clustering (\texttt{DeepC}, \texttt{SeLa}, \texttt{Swav}), contrastive learning (\texttt{SimCLR}, \texttt{MoCo}, \texttt{PCL}), distillation (\texttt{BYOL}, \texttt{SimSiam}, \texttt{OBoW}, \texttt{DINO}), information-maximization (\texttt{Barlow}, \texttt{VicReg}). As a result our experiments, we can conclude that claiming one methodology to be superior to others would be invalid.
\end{itemize}

In essence, our experiments demonstrated that the complexity of SSL performance, when evaluated only with ImageNet, is not captured correctly. As such, for newly proposed SSL frameworks, we call for the use of a wide range of analysis techniques that involve evaluations on a variety of test sets to capture distinct characteristics of different SSL approaches.


\bibliographystyle{ieeetr}
\bibliography{conference_101719}

\begin{thebibliography}{10}

\bibitem{Alexnet}
A.~Krizhevsky, I.~Sutskever, and G.~E. Hinton, ``{ImageNet Classification with Deep Convolutional Neural Networks},'' in {\em Advances in Neural Information Processing Systems}, 2012.

\bibitem{VGG}
K.~Simonyan and A.~Zisserman, ``{Very Deep Convolutional Networks For Large-Scale Image Recognition},'' {\em International Conference on Learning Representations}, 2015.

\bibitem{resnext}
S.~Xie, R.~Girshick, P.~Doll{\'a}r, Z.~Tu, and K.~He, ``{Aggregated Residual Transformations for Deep Neural Networks},'' in {\em Proceedings of the IEEE Conference on Computer Vision and Pattern Recognition}, pp.~1492--1500, 2017.

\bibitem{labeling_hard_1}
T.~Fredriksson, D.~I. Mattos, J.~Bosch, and H.~H. Olsson, ``{Data Labeling: An Empirical Investigation Into Industrial Challenges and Mitigation Strategies},'' in {\em International Conference on Product-Focused Software Process Improvement}, 2020.

\bibitem{wang2020generalizing}
Y.~Wang, Q.~Yao, J.~T. Kwok, and L.~M. Ni, ``{Generalizing From a Few Examples: A Survey on Few-shot Learning},'' {\em ACM Computing Surveys}, 2020.

\bibitem{transfer_learning}
K.~Weiss, T.~M. Khoshgoftaar, and D.~Wang, ``{A Survey of Transfer Learning},'' {\em Journal of Big data}, 2016.

\bibitem{pytorch-whitepaper}
A.~Paszke, S.~Gross, S.~Chintala, G.~Chanan, E.~Yang, Z.~DeVito, Z.~Lin, A.~Desmaison, L.~Antiga, and A.~Lerer, ``{Automatic Differentiation in PyTorch},'' 2017.

\bibitem{ILSVRC15:rus}
O.~Russakovsky, J.~Deng, H.~Su, J.~Krause, S.~Satheesh, S.~Ma, Z.~Huang, A.~Karpathy, A.~Khosla, M.~Bernstein, A.~C. Berg, and L.~Fei-Fei, ``{ImageNet Large Scale Visual Recognition Challenge},'' {\em International Journal of Computer Vision}, vol.~115, no.~3, pp.~211--252, 2015.

\bibitem{Simclr}
T.~Chen, S.~Kornblith, M.~Norouzi, and G.~Hinton, ``{A Simple Framework for Contrastive Learning of Visual Representations},'' in {\em International Conference on Machine Learning}, 2020.

\bibitem{BYOL}
J.-B. Grill, F.~Strub, F.~Altch{\'e}, C.~Tallec, P.~Richemond, E.~Buchatskaya, C.~Doersch, B.~Avila~Pires, Z.~Guo, M.~Gheshlaghi~Azar, {\em et~al.}, ``Bootstrap your own latent-a new approach to self-supervised learning,'' {\em Advances in Neural Information Processing Systems}, 2020.

\bibitem{Mocov1}
K.~He, H.~Fan, Y.~Wu, S.~Xie, and R.~Girshick, ``{Momentum Contrast for Unsupervised Visual Representation Learning},'' in {\em Proceedings of the IEEE/CVF Conference on Computer Vision and Pattern Recognition}, 2020.

\bibitem{vicreg}
A.~Bardes, J.~Ponce, and Y.~LeCun, ``{Vicreg: Variance-Invariance-Covariance Regularization for Self-supervised Learning},'' {\em arXiv preprint arXiv:2105.04906}, 2021.

\bibitem{barlow}
J.~Zbontar, L.~Jing, I.~Misra, Y.~LeCun, and S.~Deny, ``{Barlow Twins: Self-supervised Learning via Redundancy Reduction},'' in {\em International Conference on Machine Learning}, 2021.

\bibitem{pcl}
J.~Li, P.~Zhou, C.~Xiong, and S.~Hoi, ``{Prototypical Contrastive Learning of Unsupervised Representations},'' in {\em International Conference on Learning Representations}, 2021.

\bibitem{Swav}
M.~Caron, I.~Misra, J.~Mairal, P.~Goyal, P.~Bojanowski, and A.~Joulin, ``Unsupervised learning of visual features by contrasting cluster assignments,'' {\em Advances in Neural Information Processing Systems}, 2020.

\bibitem{Pirl}
I.~Misra and L.~v.~d. Maaten, ``{Self-supervised Learning of Pretext-invariant Representations},'' in {\em Proceedings of the IEEE/CVF Conference on Computer Vision and Pattern Recognition}, 2020.

\bibitem{2023-balestriero-CookbookSelfSupervisedLearning}
R.~Balestriero, M.~Ibrahim, V.~Sobal, A.~Morcos, S.~Shekhar, T.~Goldstein, F.~Bordes, A.~Bardes, G.~Mialon, Y.~Tian, A.~Schwarzschild, A.~G. Wilson, J.~Geiping, Q.~Garrido, P.~Fernandez, A.~Bar, H.~Pirsiavash, Y.~LeCun, and M.~Goldblum, ``A {{Cookbook}} of {{Self-Supervised Learning}},'' {\em arXiv preprint arXiv:2304.12210}, 2023.

\bibitem{2023-ozbulak-KnowYourSelfsupervised}
U.~Ozbulak, H.~J. Lee, B.~Boga, E.~T. Anzaku, H.-m. Park, A.~V. Messem, W.~D. Neve, and J.~Vankerschaver, ``Know {{Your Self-supervised Learning}}: {{A Survey}} on {{Image-based Generative}} and {{Discriminative Training}},'' {\em Transactions on Machine Learning Research}, 2023.

\bibitem{musgrave2020metric}
K.~Musgrave, S.~Belongie, and S.-N. Lim, ``A metric learning reality check,'' in {\em Computer Vision--ECCV 2020: 16th European Conference, Glasgow, UK, August 23--28, 2020, Proceedings, Part XXV 16}, pp.~681--699, Springer, 2020.

\bibitem{dehghani2021benchmark}
M.~Dehghani, Y.~Tay, A.~A. Gritsenko, Z.~Zhao, N.~Houlsby, F.~Diaz, D.~Metzler, and O.~Vinyals, ``The benchmark lottery,'' {\em arXiv preprint arXiv:2107.07002}, 2021.

\bibitem{Mocov3}
X.~Chen, S.~Xie, and K.~He, ``{An Empirical Study of Training Self-supervised Vision Transformers},'' in {\em Proceedings of the IEEE/CVF International Conference on Computer Vision}, 2021.

\bibitem{becker1992self}
S.~Becker and G.~E. Hinton, ``Self-organizing neural network that discovers surfaces in random-dot stereograms,'' {\em Nature}, 1992.

\bibitem{de1994learning}
V.~R. de~Sa, ``Learning classification with unlabeled data,'' {\em Advances in Neural Information Processing Systems}, 1994.

\bibitem{image_colorization}
R.~Zhang, P.~Isola, and A.~A. Efros, ``{Colorful Image Colorization},'' in {\em European Conference on Computer Vision}, 2016.

\bibitem{image_colorization2}
G.~Larsson, M.~Maire, and G.~Shakhnarovich, ``{Learning Representations for Automatic Colorization},'' in {\em European Conference on Computer Vision}, 2016.

\bibitem{super_resolution}
C.~Ledig, L.~Theis, F.~Husz{\'a}r, J.~Caballero, A.~Cunningham, A.~Acosta, A.~Aitken, A.~Tejani, J.~Totz, Z.~Wang, {\em et~al.}, ``Photo-realistic single image super-resolution using a generative adversarial network,'' in {\em Proceedings of the IEEE Conference on Computer Vision and Pattern Recognition}, 2017.

\bibitem{inpainting}
D.~Pathak, P.~Krahenbuhl, J.~Donahue, T.~Darrell, and A.~A. Efros, ``Context encoders: Feature learning by inpainting,'' in {\em Proceedings of the IEEE conference on computer vision and pattern recognition}, pp.~2536--2544, 2016.

\bibitem{geometric_trans}
S.~Gidaris, P.~Singh, and N.~Komodakis, ``{Unsupervised Representation Learning by Predicting Image Rotations},'' {\em arXiv preprint arXiv:1803.07728}, 2018.

\bibitem{ssl_early}
C.~Doersch, A.~Gupta, and A.~A. Efros, ``{Unsupervised Visual Representation Learning by Context Prediction},'' in {\em Proceedings of the IEEE International Conference on Computer Vision}, 2015.

\bibitem{ssl_early2}
R.~Zhang, P.~Isola, and A.~A. Efros, ``{Split-brain Autoencoders: Unsupervised Learning by Cross-channel Prediction},'' in {\em Proceedings of the IEEE Conference on Computer Vision and Pattern Recognition}, 2017.

\bibitem{deep_cluster}
M.~Caron, P.~Bojanowski, A.~Joulin, and M.~Douze, ``{Deep Clustering for Unsupervised Learning of Visual Features},'' in {\em European Conference on Computer Vision}, 2018.

\bibitem{sela}
Y.~M. Asano, C.~Rupprecht, and A.~Vedaldi, ``{Self-labelling via Simultaneous Clustering and Representation Learning},'' {\em arXiv preprint arXiv:1911.05371}, 2019.

\bibitem{Obow}
S.~Gidaris, A.~Bursuc, G.~Puy, N.~Komodakis, M.~Cord, and P.~Perez, ``Obow: Online bag-of-visual-words generation for self-supervised learning,'' in {\em Proceedings of the IEEE/CVF Conference on Computer Vision and Pattern Recognition}, 2021.

\bibitem{Simsiam}
X.~Chen and K.~He, ``{Exploring Simple Siamese Representation Learning},'' in {\em Proceedings of the IEEE/CVF Conference on Computer Vision and Pattern Recognition}, 2021.

\bibitem{Dino}
M.~Caron, H.~Touvron, I.~Misra, H.~J{\'e}gou, J.~Mairal, P.~Bojanowski, and A.~Joulin, ``Emerging properties in self-supervised vision transformers,'' in {\em Proceedings of the IEEE/CVF International Conference on Computer Vision}, 2021.

\bibitem{feature_quantization}
J.~Johnson, M.~Douze, and H.~J{\'e}gou, ``Billion-scale similarity search with gpus,'' {\em IEEE Transactions on Big Data}, 2019.

\bibitem{infonce_loss}
A.~v.~d. Oord, Y.~Li, and O.~Vinyals, ``{Representation Learning with Contrastive Predictive Coding},'' {\em arXiv preprint arXiv:1807.03748}, 2018.

\bibitem{gidaris2020learning}
S.~Gidaris, A.~Bursuc, N.~Komodakis, P.~P{\'e}rez, and M.~Cord, ``{Learning Representations by Predicting Bags of Visual Words},'' in {\em Proceedings of the IEEE/CVF Conference on Computer Vision and Pattern Recognition}, 2020.

\bibitem{lecun1998gradient}
Y.~LeCun, L.~Bottou, Y.~Bengio, and P.~Haffner, ``{Gradient-Based Learning Applied To Document Recognition},'' {\em Proceedings of the IEEE}, 1998.

\bibitem{coco}
T.-Y. Lin, M.~Maire, S.~Belongie, J.~Hays, P.~Perona, D.~Ramanan, P.~Doll{\'a}r, and C.~L. Zitnick, ``{Microsoft Coco: Common Objects In Context},'' in {\em Proceedings of the IEEE European Conference on Computer Vision}, pp.~740--755, Springer, 2014.

\bibitem{zhai2019large}
X.~Zhai, J.~Puigcerver, A.~Kolesnikov, P.~Ruyssen, C.~Riquelme, M.~Lucic, J.~Djolonga, A.~S. Pinto, M.~Neumann, A.~Dosovitskiy, {\em et~al.}, ``{A Large-scale Study of Representation Learning with the Visual Task Adaptation Benchmark},'' {\em arXiv preprint arXiv:1910.04867}, 2019.

\bibitem{goyal2019scaling}
P.~Goyal, D.~Mahajan, A.~Gupta, and I.~Misra, ``{Scaling and Benchmarking Self-supervised Visual Representation Learning},'' in {\em Proceedings of the ieee/cvf International Conference on computer vision}, 2019.

\bibitem{kornblith2019better}
S.~Kornblith, J.~Shlens, and Q.~V. Le, ``{Do Better Imagenet Models Transfer Better?},'' in {\em Proceedings of the IEEE/CVF conference on computer vision and pattern recognition}, 2019.

\bibitem{how_well_transfer}
L.~Ericsson, H.~Gouk, and T.~M. Hospedales, ``{How Well Do Self-supervised Models Transfer?},'' in {\em Proceedings of the IEEE/CVF Conference on Computer Vision and Pattern Recognition}, 2021.

\bibitem{imagenet_a_o}
D.~Hendrycks, K.~Zhao, S.~Basart, J.~Steinhardt, and D.~Song, ``{Natural Adversarial Examples},'' in {\em Proceedings of the IEEE/CVF Conference on Computer Vision and Pattern Recognition}, 2021.

\bibitem{imagenet_sketch}
H.~Wang, S.~Ge, Z.~Lipton, and E.~P. Xing, ``{Learning Robust Global Representations by Penalizing Local Predictive Power},'' {\em Advances in Neural Information Processing Systems}, 2019.

\bibitem{imagenet_r}
D.~Hendrycks, S.~Basart, N.~Mu, S.~Kadavath, F.~Wang, E.~Dorundo, R.~Desai, T.~Zhu, S.~Parajuli, M.~Guo, {\em et~al.}, ``{The many faces of robustness: A critical analysis of out-of-distribution generalization},'' in {\em Proceedings of the IEEE/CVF International Conference on Computer Vision}, 2021.

\bibitem{imagenet_r_evaluation}
S.~Schneider, E.~Rusak, L.~Eck, O.~Bringmann, W.~Brendel, and M.~Bethge, ``{Improving Robustness Against Common Corruptions by Covariate Shift Adaptation},'' {\em Advances in Neural Information Processing Systems}, vol.~33, pp.~11539--11551, 2020.

\bibitem{imagenetv2}
B.~Recht, R.~Roelofs, L.~Schmidt, and V.~Shankar, ``Do imagenet classifiers generalize to imagenet?,'' in {\em International Conference on Machine Learning}, 2019.

\bibitem{zhang2020impact}
Y.~Zhang and B.~D. Davison, ``{Impact of ImageNet Model Selection on Domain Adaptation},'' in {\em Proceedings of the IEEE/CVF Winter Conference on Applications of Computer Vision Workshops}, 2020.

\bibitem{zhang2021semi}
Y.~Zhang, H.~Zhang, B.~Deng, S.~Li, K.~Jia, and L.~Zhang, ``{Semi-supervised Models are Strong Unsupervised Domain Adaptation Learners},'' {\em arXiv preprint arXiv:2106.00417}, 2021.

\bibitem{hendrycks2016baseline}
D.~Hendrycks and K.~Gimpel, ``{A Baseline for Detecting Misclassified and Out-of-distribution Examples in Neural Networks},'' {\em arXiv preprint arXiv:1610.02136}, 2016.

\bibitem{winkens2020contrastive}
J.~Winkens, R.~Bunel, A.~G. Roy, R.~Stanforth, V.~Natarajan, J.~R. Ledsam, P.~MacWilliams, P.~Kohli, A.~Karthikesalingam, S.~Kohl, {\em et~al.}, ``{Contrastive Training for Improved Out-of-distribution Detection},'' {\em arXiv preprint arXiv:2007.05566}, 2020.

\bibitem{ozbulak2021evaluating}
U.~Ozbulak, M.~Pintor, A.~Van~Messem, and W.~De~Neve, ``Evaluating adversarial attacks on imagenet: A reality check on misclassification classes,'' {\em arXiv preprint arXiv:2111.11056}, 2021.

\bibitem{resnet}
K.~He, X.~Zhang, S.~Ren, and J.~Sun, ``{Deep Residual Learning For Image Recognition},'' in {\em Proceedings of the IEEE Conference on Computer Vision and Pattern Recognition}, 2016.

\bibitem{Inceptionv1}
C.~Szegedy, W.~Liu, Y.~Jia, P.~Sermanet, S.~Reed, D.~Anguelov, D.~Erhan, V.~Vanhoucke, and A.~Rabinovich, ``{Going Deeper With Convolutions},'' in {\em Proceedings of the IEEE Conference on Computer Vision and Pattern Recognition}, pp.~1--9, 2015.

\bibitem{imagenet_problem1}
C.~Northcutt, L.~Jiang, and I.~Chuang, ``{Confident Learning: Estimating Uncertainty in Dataset Labels},'' {\em Journal of Artificial Intelligence Research}, 2021.

\bibitem{imagenet_problem2}
S.~Hooker, Y.~Dauphin, A.~Courville, and A.~Frome, ``Selective brain damage: Measuring the disparate impact of model pruning,'' 2019.

\bibitem{ImagenetReal}
L.~Beyer, O.~J. H{\'e}naff, A.~Kolesnikov, X.~Zhai, and A.~v.~d. Oord, ``{Are We Done with Imagenet?},'' {\em arXiv preprint arXiv:2006.07159}, 2020.

\bibitem{dnn_texture_bias}
R.~Geirhos, P.~Rubisch, C.~Michaelis, M.~Bethge, F.~A. Wichmann, and W.~Brendel, ``{ImageNet-trained CNNs are Biased Towards Texture; Increasing Shape Bias Improves Accuracy and Robustness},'' {\em arXiv preprint arXiv:1811.12231}, 2018.

\bibitem{Goodfellow-expharnessing}
I.~Goodfellow, J.~Shlens, and C.~Szegedy, ``{Explaining and Harnessing Adversarial Examples},'' {\em International Conference on Learning Representations}, 2015.

\bibitem{IFGS}
A.~Kurakin, I.~Goodfellow, and S.~Bengio, ``{Adversarial Examples In The Physical World},'' {\em Workshop Track, International Conference on Learning Representations}, 2016.

\bibitem{2009-borenstein-IntroductionMetaAnalysis}
M.~Borenstein, L.~V. Hedges, J.~P.~T. Higgins, and H.~R. Rothstein, {\em Introduction to {{Meta-Analysis}}}.
\newblock {Wiley}, 2009.

\end{thebibliography}

\end{document}